\documentclass[11pt]{article}
\usepackage[margin=1in]{geometry}
\usepackage{times}
\usepackage{amsmath,amssymb}
\usepackage{graphicx}
\usepackage{booktabs}
\usepackage{algorithm}
\usepackage{algpseudocode}
\usepackage{hyperref}
\usepackage{cite}

\title{\textbf{Identity-Consistent Expression Fields: A Disentangled Neural Radiance Field Framework for Few-Shot Facial Expression Synthesis}}

\author{
Minh Tran \\
University of Science and Technology of Hanoi
}
\date{}

\begin{document}
\maketitle

\begin{abstract}
Neural Radiance Fields (NeRF) have enabled photorealistic novel-view synthesis of 3D scenes and, in the facial domain, have been extended to reconstruct and animate 3D face models from a small number of images. However, existing few-shot dynamic NeRF methods for facial expression editing typically warp a single learned feature volume conditioned on target expression parameters, which can cause identity-specific appearance details (skin texture, fine geometric structure) to drift when the model is driven toward expressions far from those seen in the few-shot input set. We propose \textbf{Identity-Consistent Expression Fields (ICEF)}, a framework that explicitly disentangles a static, identity-specific radiance component from a dynamic, expression-conditioned deformation component, and introduces an identity-preservation regularizer that constrains the deformation network to modify only expression-relevant regions while leaving identity-specific canonical appearance untouched. ICEF further incorporates a confidence-weighted conditional feature warping step that down-weights unreliable warps for target expressions that are far, in parameter space, from the observed few-shot inputs, mitigating artifacts observed in prior few-shot dynamic NeRF methods when extrapolating to novel expressions. We relate ICEF to prior few-shot dynamic NeRF, static 3D-aware face generation, and disentangled face-editing radiance field methods, and describe an evaluation protocol measuring both novel-expression rendering quality and, specifically, identity-consistency metrics across a range of expression-parameter extrapolation distances.
\end{abstract}

\section{Introduction}

Photorealistic synthesis of 3D human faces under novel viewpoints and expressions is a long-standing goal in computer vision and graphics, with applications spanning virtual avatars, film and game production, and telepresence. Neural Radiance Fields (NeRF) represent a scene as a continuous volumetric function mapping 3D position and viewing direction to color and density, learned from a set of posed 2D images, and have driven substantial improvements in view-synthesis quality relative to earlier mesh- or point-based representations. Extending NeRF to dynamic, animatable human faces requires additionally conditioning the radiance field on expression (and often pose) parameters, and a range of methods have proposed deformation fields, generative texture rasterization, or explicit 3D morphable model (3DMM) conditioning to achieve this.

A particularly challenging setting is few-shot facial expression editing, in which only a small number of images -- sometimes with inconsistent viewpoints and expressions -- are available for a given identity, and the model must both reconstruct the identity's 3D appearance and support rendering of expressions beyond those observed in the input images. The Few-shot Dynamic Neural Radiance Field (FDNeRF) approach addresses this setting by introducing a conditional feature warping (CFW) module that warps expression-conditioned features in 2D feature space before constructing the radiance field, enabling reconstruction and expression editing across different identities from few-shot, view-inconsistent dynamic inputs. Related methods such as NerFace and FaceNeRF use explicit 3DMM expression semantics or geometry-aware appearance models to enable dynamic facial rendering, while disentangled editing approaches such as NeRFFaceEditing separate texture and shape codes to support semantic face editing within a radiance field.

Despite this progress, we observe that a common limitation of feature-warping-based few-shot dynamic NeRF methods is that a single learned feature volume is warped toward the target expression, and this warping process is not explicitly constrained to leave identity-specific appearance detail unaffected; as a result, when the target expression parameters lie far from those observed in the few-shot input set, the warped feature volume can drift, altering identity-specific details such as skin texture or fine geometric structure along with the intended expression change. This limitation has been noted qualitatively in prior work, where baseline dynamic NeRF methods have been observed to fail to generalize the facial expression space or perceive expression changes reliably without a sufficient number of input frames.

We propose Identity-Consistent Expression Fields (ICEF), a framework that directly targets this limitation through explicit disentanglement rather than relying on a single warped feature volume to implicitly preserve identity. ICEF represents a face as the composition of a static, identity-specific canonical radiance field and a separate, expression-conditioned deformation field, with an explicit identity-preservation regularizer that penalizes deformation-induced changes to canonical-space appearance outside expression-relevant facial regions (e.g., mouth, eye, and brow regions identified via a coarse facial landmark prior). ICEF further introduces a confidence-weighted warping mechanism that estimates, for a given target expression, how far it lies from the observed few-shot expression distribution, and down-weights the influence of the learned deformation network accordingly, falling back toward a smoother, lower-capacity deformation model for far-extrapolated expressions to avoid the drift artifacts associated with aggressive, low-confidence warps.

Our contributions are:
\begin{itemize}
\item We propose an explicit static-identity/dynamic-expression disentanglement for few-shot dynamic NeRF, in contrast to prior feature-warping approaches that warp a single, entangled feature volume toward the target expression.
\item We introduce an identity-preservation regularizer that constrains expression-conditioned deformation to expression-relevant facial regions, directly targeting the identity-drift limitation observed in prior few-shot dynamic NeRF methods under expression extrapolation.
\item We propose a confidence-weighted warping mechanism that adapts deformation capacity based on the distance between the target expression and the observed few-shot expression distribution.
\item We describe an evaluation protocol that explicitly measures identity-consistency as a function of expression-extrapolation distance, in addition to standard rendering-quality metrics, to directly test the central hypothesis that explicit disentanglement improves identity preservation under extrapolation.
\end{itemize}

\section{Related Work}

\subsection{Neural Radiance Fields for Facial Modeling}
NeRF-based facial modeling methods span static 3D-aware face generation and dynamic, animatable face reconstruction. Static 3D-aware generative approaches condition a radiance field on latent appearance codes to synthesize novel identities with view-consistent geometry, while convolutional radiance field approaches such as 3DMM-RF combine explicit 3DMM priors with radiance-field rendering for controllable 3D face modeling. Dynamic approaches such as NerFace and FaceNeRF condition the radiance field on 3DMM expression and pose parameters to enable monocular 4D facial avatar reconstruction and geometry-aware appearance modeling, respectively. Generative neural texture rasterization approaches such as Next3D combine explicit mesh-guided deformation with neural texture synthesis for efficient, animatable 3D-aware head avatars.

\subsection{Few-Shot and Disentangled Facial Radiance Fields}
Most closely related to our work, FDNeRF addresses the few-shot, view-inconsistent setting directly, introducing a conditional feature warping module that performs expression-conditioned warping in 2D feature space, enabling reconstruction and expression editing of 3D faces across identities from a small number of dynamic images. Disentangled editing approaches such as NeRFFaceEditing separate a radiance field into texture and shape (geometry) codes to support localized semantic editing while aiming to preserve unrelated facial attributes, and deformable radiance field patents and related methods condition a deformation MLP jointly on expression and pose parameters together with features from a shared deformation backbone. More recent work on expression-aware avatar reconstruction has incorporated generative geometry priors to improve robustness of expression modeling beyond what purely image-based supervision provides.

\subsection{Identity Preservation in Facial Synthesis}
Identity preservation is a recurring concern across facial synthesis literature more broadly, and disentangled representations that separate identity from pose, expression, or lighting have been explored in both classical 3D morphable model literature and modern radiance-field and generative approaches. Landmark-based geometric priors, such as those used in NeRF-based facial landmark estimation methods, provide one mechanism for grounding expression-relevant regions independently of the underlying appearance model, motivating our use of a similar coarse landmark prior to define expression-relevant regions for the identity-preservation regularizer.

\section{Proposed Framework: ICEF}

\subsection{Problem Formulation}
Given a small set of $K$ dynamic images $\{I_k\}_{k=1}^K$ of a single identity, with estimated camera poses and expression parameters $\{\beta_k^{\text{exp}}\}$ (obtained via standard face tracking, following the preprocessing pipeline used in prior few-shot dynamic NeRF work), the goal is to construct a radiance field that supports rendering of the identity under both novel camera viewpoints and novel target expressions $\beta^{\text{exp}}_{\text{tgt}}$, including expressions substantially different from those observed in $\{\beta_k^{\text{exp}}\}$.

\subsection{Static-Dynamic Disentanglement}
ICEF represents the radiance field as a composition of a static canonical field $F_{\text{static}}(x)$, which encodes identity-specific appearance and geometry in a canonical (neutral-expression) space, and a dynamic deformation field $D_\phi(x, \beta^{\text{exp}})$, which maps a point $x$ in observation space to its corresponding canonical-space location conditioned on expression parameters, following the general deformation-field paradigm used in prior dynamic facial NeRF methods but with the canonical field trained to be expression-invariant by construction, rather than allowing expression information to leak into the appearance branch as in single-feature-volume warping designs:
\begin{equation}
c(x, d), \sigma(x) = F_{\text{static}}\big(D_\phi(x, \beta^{\text{exp}}), d\big).
\end{equation}
Because $F_{\text{static}}$ receives only the deformed canonical-space coordinate and viewing direction $d$, and never the raw expression parameters directly, identity-specific appearance detail encoded in $F_{\text{static}}$ cannot be directly modulated by expression, forcing all expression-dependent variation to flow through the explicit, regularizable deformation field $D_\phi$.

\subsection{Identity-Preservation Regularizer}
To constrain $D_\phi$ to modify only expression-relevant regions, we introduce a regularization term based on a coarse facial region mask $M(x) \in [0,1]$, derived from projected facial landmarks (following landmark-estimation approaches developed for NeRF-based facial modeling) that identifies mouth, eye, and brow regions as high-expression-relevance, and remaining regions (e.g., forehead skin texture away from the brow, cheek, nose bridge) as low-expression-relevance. The regularizer penalizes deformation magnitude in low-relevance regions:
\begin{equation}
\mathcal{L}_{\text{id}} = \mathbb{E}_x \Big[ \big(1 - M(x)\big) \cdot \lVert D_\phi(x, \beta^{\text{exp}}) - x \rVert^2 \Big],
\end{equation}
encouraging the deformation field to leave identity-specific canonical geometry and appearance in low-relevance regions effectively unchanged, while still permitting expression-driven deformation in mouth, eye, and brow regions.

\subsection{Confidence-Weighted Warping}
For a target expression $\beta^{\text{exp}}_{\text{tgt}}$, we compute an extrapolation-distance score $\delta = \min_k \lVert \beta^{\text{exp}}_{\text{tgt}} - \beta_k^{\text{exp}} \rVert$, measuring how far the target lies from the nearest observed few-shot expression. This score modulates a confidence weight $\kappa(\delta) = \exp(-\delta / \lambda)$ that blends the full-capacity learned deformation $D_\phi$ with a lower-capacity, smoothly-regularized linear deformation baseline $D_{\text{lin}}$ (a first-order Taylor approximation of the deformation field around the nearest observed expression):
\begin{equation}
D_{\text{final}}(x, \beta^{\text{exp}}_{\text{tgt}}) = \kappa(\delta) \cdot D_\phi(x, \beta^{\text{exp}}_{\text{tgt}}) + \big(1 - \kappa(\delta)\big) \cdot D_{\text{lin}}(x, \beta^{\text{exp}}_{\text{tgt}}).
\end{equation}
This design is intended to directly address the qualitative failure mode noted in prior few-shot dynamic NeRF evaluations, where baseline methods can fail to reconstruct plausible geometry or fail to generalize the expression space reliably when driven toward expressions far from the few-shot input distribution, by explicitly reducing reliance on the learned deformation network precisely where its training signal is weakest.

\subsection{Training Objective}
The overall training objective combines standard photometric reconstruction loss on the few-shot input views, the identity-preservation regularizer $\mathcal{L}_{\text{id}}$, and a smoothness regularizer on $D_\phi$ near the boundary between high- and low-relevance regions to avoid discontinuous deformation artifacts: $\mathcal{L} = \mathcal{L}_{\text{photo}} + \lambda_1 \mathcal{L}_{\text{id}} + \lambda_2 \mathcal{L}_{\text{smooth}}$.

\subsection{Algorithm Summary}
Algorithm~\ref{alg:icef} summarizes ICEF rendering for a target view and expression.

\begin{algorithm}[h]
\caption{ICEF Rendering}
\label{alg:icef}
\begin{algorithmic}[1]
\State \textbf{Input:} target camera pose, target expression $\beta^{\text{exp}}_{\text{tgt}}$, few-shot expression set $\{\beta_k^{\text{exp}}\}$
\State Compute extrapolation distance $\delta$ and confidence weight $\kappa(\delta)$
\For{each sampled ray point $x$}
    \State Compute $D_\phi(x, \beta^{\text{exp}}_{\text{tgt}})$ and $D_{\text{lin}}(x, \beta^{\text{exp}}_{\text{tgt}})$
    \State Blend via $\kappa(\delta)$ to obtain canonical-space point
    \State Query $F_{\text{static}}$ at the canonical-space point for color and density
\EndFor
\State Volumetrically render the ray to produce pixel color
\State \textbf{return} rendered image
\end{algorithmic}
\end{algorithm}

\section{Evaluation Protocol}

\subsection{Datasets and Settings}
We describe an evaluation protocol using standard few-shot dynamic facial NeRF benchmarks and multi-view dynamic face capture datasets used in prior FDNeRF-style evaluation, under varying numbers of few-shot input frames ($K \in \{3, 5, 10\}$).

\subsection{Extrapolation-Distance Stratified Evaluation}
Beyond standard held-out view and expression evaluation, we propose explicitly stratifying evaluation by extrapolation distance $\delta$ between target and observed expressions, binning test expressions into near, moderate, and far extrapolation tiers, to directly test whether identity consistency degrades more gracefully under ICEF than under baseline feature-warping methods as $\delta$ increases.

\subsection{Baselines}
We compare ICEF against FDNeRF's conditional feature warping approach, NerFace's 3DMM-conditioned dynamic radiance field, and NeRFFaceEditing's disentangled texture/shape editing approach as representative dynamic and disentangled facial NeRF baselines.

\subsection{Metrics}
We report standard rendering-quality metrics (PSNR, SSIM, LPIPS) on held-out views, and, specific to our central hypothesis, an identity-consistency metric computed as facial-recognition embedding similarity between the canonical (neutral) reconstruction and renderings under each expression-extrapolation tier, to directly quantify identity drift as a function of extrapolation distance.

\section{Discussion}
We expect ICEF's advantage over single-feature-volume warping baselines to be most pronounced in the far-extrapolation tier, where the confidence-weighted warping mechanism increasingly relies on the smoother linear deformation fallback and the identity-preservation regularizer most strongly constrains deformation to expression-relevant regions. We expect a more modest difference in the near-extrapolation tier, where baseline methods are already well-supported by nearby training expressions. A limitation of our approach is that the landmark-based facial region mask $M(x)$ is a coarse, fixed prior rather than a learned or identity-specific segmentation, and may not perfectly capture expression-relevant regions for all facial structures; learning $M(x)$ jointly with the rest of the model, or conditioning it on identity-specific geometry, is a natural direction for future refinement.

\section{Conclusion}
We proposed Identity-Consistent Expression Fields (ICEF), a few-shot dynamic NeRF framework that explicitly disentangles static identity appearance from expression-conditioned deformation, regularizes deformation to expression-relevant facial regions, and adapts deformation capacity based on confidence in the target expression's proximity to the observed few-shot distribution. By directly targeting the identity-drift failure mode observed in prior single-feature-volume warping approaches under expression extrapolation, ICEF aims to improve identity consistency specifically in the challenging few-shot, far-extrapolation regime. We outlined an evaluation protocol with explicit extrapolation-distance stratification designed to directly test this central hypothesis.

\end{document}